\pdfoutput=1
\documentclass[11pt]{article}

\usepackage{acl}
\usepackage{latexsym, graphicx, float, amsmath, tabularx, subcaption}
\usepackage{booktabs}
\usepackage{threeparttable} % to use table notes
\usepackage{times}
\usepackage{hyperref}
\usepackage{url}
\usepackage{multirow}
\usepackage{xcolor} % import these two packages for colorful table
\usepackage{colortbl} % import these two packages for colorful table
\usepackage{multicol}
\usepackage{titlesec}
\usepackage{arydshln} 
\usepackage{inconsolata}
\usepackage{microtype}
\usepackage{amsfonts,amssymb}

\definecolor{grey}{gray}{0.5}

\usepackage[T1]{fontenc}
\usepackage[utf8]{inputenc}
\usepackage{soul}
\title{Does Table Source Matter? Benchmarking and Improving Multimodal Scientific Table Understanding and Reasoning}

\author{
Bohao Yang\textsuperscript{1},
Yingji Zhang\textsuperscript{1},
Dong Liu\textsuperscript{2}, 
Andr\'{e} Freitas\textsuperscript{1,3}, 
Chenghua Lin\textsuperscript{1}\thanks{\quad \small Corresponding author}\space \\
\textsuperscript{1} The University of Manchester
\textsuperscript{2} Tencent Timi Studio
\textsuperscript{3} Idiap Research Institute \\
\texttt{
\{bohao.yang-2, yingji.zhang\}@postgrad.manchester.ac.uk
}\vspace{-0.1mm}\\
\texttt{
\{andre.freitas, chenghua.lin\}@manchester.ac.uk,
}\vspace{-0.5mm}\\
\texttt{
dougliu@tencent.com
}\vspace{-0.5mm} 
}

\begin{document}
\maketitle
\begin{abstract}

Recent large language models (LLMs) have advanced table understanding capabilities but rely on converting tables into text sequences. While multimodal large language models (MLLMs) enable direct visual processing, they face limitations in handling scientific tables due to fixed input image resolutions and insufficient numerical reasoning capabilities.
To address these challenges, we present MMSci, a comprehensive dataset for scientific table understanding and reasoning. MMSci consists of three key components: (1) MMSci-Pre, a domain-specific dataset of 52K scientific table structure recognition samples, (2) MMSci-Ins, an instruction tuning dataset with 12K samples across three table-based tasks, and (3) MMSci-Eval, a benchmark with 3,114 testing samples specifically designed to evaluate numerical reasoning capabilities.
Based on MMSci, we develop a table-based MLLM framework with dynamic input image resolutions. Extensive experiments demonstrate that our domain-specific approach with 52K scientific table images achieves superior performance compared to 150K general-domain tables, highlighting the importance of data quality over quantity. Our proposed framework shows significant improvements in both general table understanding and numerical reasoning capabilities, with strong generalisation to held-out datasets. Our code and data are publicly available at 
\url{https://github.com/Bernard-Yang/MMSci_Table}.
% \url{https://anonymous.4open.science/r/MMSci_Table-F278/}.

% Our code and data are publicly available at \url{https://anonymous.4open.science/r/MMSci_Table-F278/}.

\end{abstract}

\section{Introduction}
\label{sec:introduction}

% Background

Tables serve as a fundamental tool for organising structured information across diverse domains. Recent studies have shown the potential of leveraging large language models (LLMs) to automatically understand and process tabular data, which has emerged as a critical research direction with applications such as Table Question Answering (TQA)~\citep{WTQ}, Table Fact Verification (TFV)~\cite{TabFact}, and Table-to-Text Generation (T2T)~\citep{moosavi2021scigen}. 

However, current table-oriented LLMs~\citep{tablellama,tablegpt_chatgpt} face inherent limitations as they require converting tables into sequential text formats (i.e., HTML strings), potentially losing crucial structural and positional information. While table-based multimodal large language models (MLLMs) have addressed this by enabling direct processing of table images, several critical limitations persist: (1) fixed input image resolutions that constrain practical applicability, (2) limited capability in processing scientific tables that contain significant numerical values, and (3) insufficient numerical reasoning abilities for scientific domain tasks.
These limitations are particularly significant in scientific domains, where tables frequently incorporate complex numerical relationships.
% , statistical analyses, and domain-specific notations. 
While current MLLMs have demonstrated efficacy with general-domain tables (e.g., tables from Wikipedia or reports), they exhibit substantial performance degradation when confronted with scientific tables that mostly contains numerical values. Processing such tables effectively requires not only visual understanding to capture structural elements but also advanced numerical reasoning capabilities to interpret and analyse complex numerical relationships, statistical significance, and experimental results. Current MLLMs, however, lack the specific architectural designs and domain-specific training data to handle these sophisticated scientific table understanding requirements.

To address these challenges, we introduce MMSci, a comprehensive dataset for scientific table understanding and reasoning. We first conduct a systematic analysis of table source effectiveness through MMSci-Pre, a carefully curated dataset containing 52K structure recognition samples derived from scientific papers. Our experimental results demonstrate that MLLMs trained on these domain-specific table images significantly outperform those trained on 150K general-domain tables, establishing the importance of data quality over quantity in table understanding tasks.
% Building upon this foundation, we develop MMSci-Ins, an instruction tuning dataset comprising 12K samples, MMSci-Eval, a benchmark with 3,114 testing samples requiring numerical reasoning for comprehensive evaluation.with explicit intermediate reasoning steps across three fundamental tasks: TQA, TFV, and T2T. Each sample incorporates detailed step-by-step reasoning processes, enabling models to develop robust mathematical reasoning and scientific analysis capabilities. 

Building upon this foundation, we then create MMSci-Ins, an instruction tuning dataset comprising 12K samples with explicit intermediate reasoning steps across three fundamental tasks: TQA, TFV, and T2T. Each sample includes detailed step-by-step reasoning processes to develop models' table-based numerical reasoning and scientific analysis capabilities.
To overcome the limitations of fixed-resolution approaches in existing table MLLMs~\cite{lee_pix2struct_2023, alonso-etal-2024-pixt3, zheng_multimodal_2024}, we implement our MLLM with dynamic input image resolution capabilities on two distinct model architectures (Qwen2-VL-7B-Instruct and LLaVA-NeXT-7B). Experimental results demonstrate consistent performance improvements across both general table understanding and specialised numerical reasoning tasks.
% Experiments demonstrate that our table-based MLLMs framework achieves superior performance with significantly less training data - specifically, our 52K scientific table images prove more effective than 150K general-domain table images for both general understanding and numerical reasoning tasks. This efficiency highlights the value of domain-specific, high-quality data in developing robust table understanding capabilities.
% Our extensive experiments demonstrate that MMSci achieves superior performance with remarkable data efficiency - our 52K scientific table images outperform 150K general-domain table images across both general understanding and specialized numerical reasoning tasks. The results show robust generalization to held-out datasets, validating the effectiveness of our approach in developing transferable table understanding capabilities. This finding highlights the critical importance of domain-specific, high-quality data and dynamic resolution processing in advancing multimodal table understanding.

To enable comprehensive evaluation, we establish MMSci-Eval, a benchmark with 3,114 testing samples requiring numerical reasoning capabilities. The benchmark provides rigorous assessment of models' performance across TQA, TFV, and T2T tasks.
% Our extensive experiments demonstrate that MMSci achieves superior performance with remarkable data efficiency. Specifically, our 52K scientific table images prove more effective than 150K general-domain table images for both general understanding and numerical reasoning tasks. This efficiency highlights the value of domain-specific, high-quality data in developing robust table understanding capabilities.
Our extensive experiments demonstrate that our 52K scientific table images prove more effective than 150K general-domain table images for both general understanding and numerical reasoning tasks. This efficiency highlights the value of domain-specific, high-quality data in developing robust table understanding capabilities.

Our contributions are summarised as follows:
\begin{itemize}
    \item We introduce MMSci, a comprehensive dataset consisting of three components: (1) MMSci-Pre, consists of 52K table image-to-HTML table structure recognition samples; (2) MMSci-Ins, an instruction tuning dataset of 12K samples with reasoning steps; and (3) MMSci-Eval, a benchmark with 3,114 samples for numerical reasoning capabilities assessment across TQA, TFV, and T2T tasks.

    \item We develop a comprehensive table-based MLLM framework that achieves strong performance on three table-based numerical reasoning tasks while demonstrating robust generalisation to held-out datasets.
    
    \item We implement dynamic input resolution capabilities across different model architectures, validating the effectiveness of our approach through consistent performance gains on both Qwen2-VL-7B-Instruct and LLaVA-NeXT-7B.
\end{itemize}

\begin{figure*}[ht]
% \small
\centering
\includegraphics[width=0.99\linewidth]{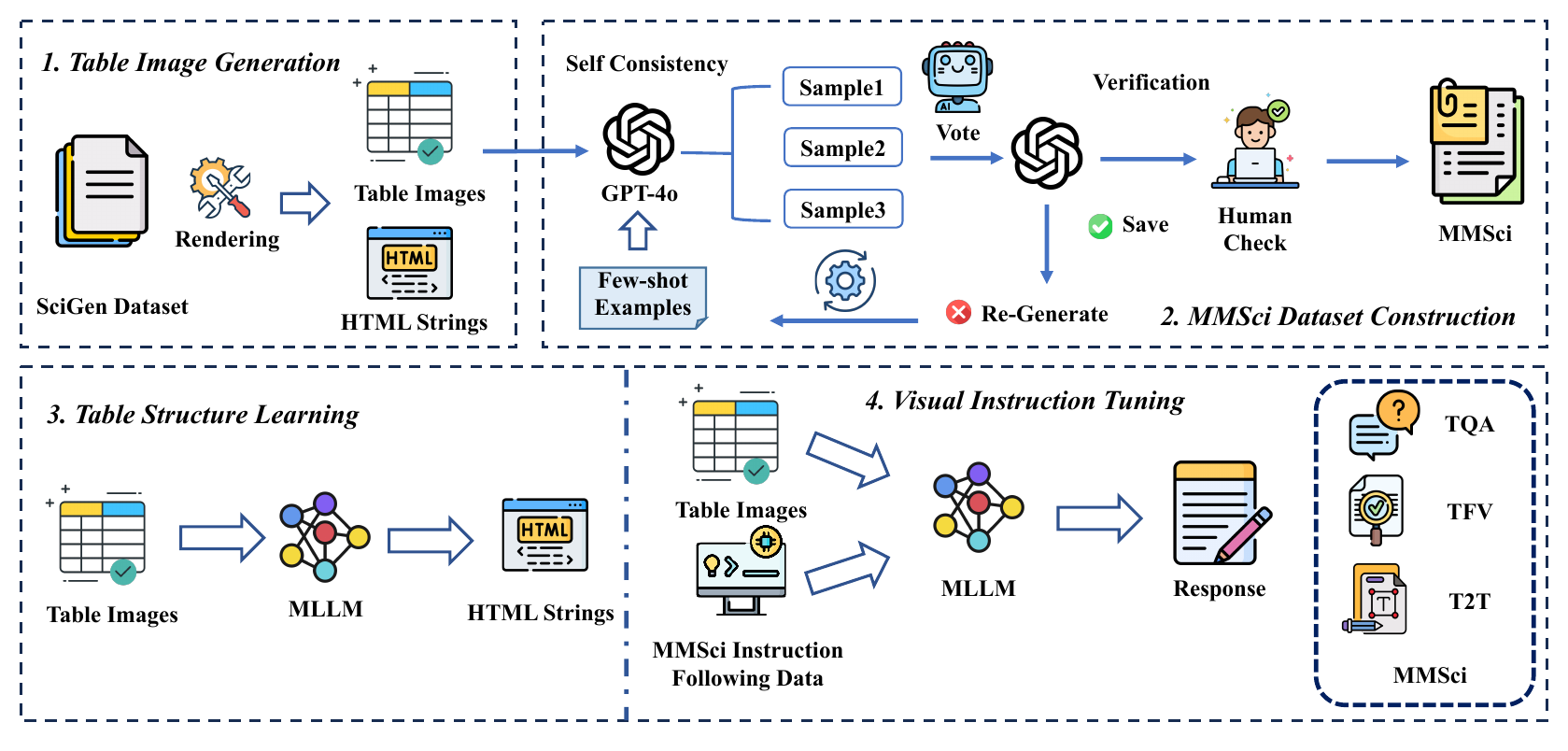}
\caption{Overview of the proposed framework, which consists of four key stages: (1) Table Image Generation; (2) MMSci Dataset Construction; (3) Table Structure Learning; and (4) Visual Instruction Tuning. }
\label{fig:model}
\end{figure*}

\section{Related Work}
\subsection{Table Understanding Models}
Early table-based models based on general language models with large-scale table corpus~\citep{liu2022tapex,hytrel} only support limited types of tables and tasks. Table understanding capabilities have been enhanced through prompt engineering~\citep{table_cot,gpt4table}, instruction tuning~\citep{tablellama,tablegpt_chatgpt,yang2024effective} and external tools~\citep{lu2023chameleon,sheetcopilot} with the development of LLMs. However, these approaches require converting tables into text formats, limiting their applications. 

Recently, MLLMs have emerged as a promising direction for table understanding. TableGPT2~\cite{su2024tablegpt2} 
% is a large multimodal model pre-trained on over 593.8K tables and 2.36M query-table-output tuples that 
features a novel table encoder to handle table cell-level information. Pix2Struct~\cite{lee_pix2struct_2023} introduces a unified image-to-text model pretrained on web page screenshots with HTML supervision. PixT3~\cite{alonso-etal-2024-pixt3} takes table-to-text tasks as table visual recognition tasks and generates texts. Table-LLaVA~\cite{zheng_multimodal_2024} introduces a novel multimodal table understanding approach that directly processes table images. However, these approaches do not focus on datasets requiring sophisticated numerical reasoning capabilities.

% TableGPT2~\cite{su2024tablegpt2} is a large multimodal model that features a novel table encoder to handle schema and cell-level information. Pix2Struct~\cite{lee_pix2struct_2023} introduces a unified image-to-text model pretrained on web page screenshots with HTML supervision, achieving state-of-the-art results across diverse visually-situated language tasks through variable-resolution inputs. PixT3~\cite{alonso-etal-2024-pixt3}is a multimodal table-to-text model that takes table-to-text tasks as table visual recognition tasks and generates texts, removing the need to process tables in text formats. Table-LLaVA~\cite{zheng_multimodal_2024} introduces a novel multimodal table understanding approach that directly processes table images rather than requiring text-based table representations.

\subsection{Table-based Reasoning and Datasets}
Table-based reasoning requires reasoning over both free-form natural language queries and structured tables. Early works either rely on executable languages (\textit{e.g.}, SQL)~\cite{yin2016neural,yu-etal-2018-spider} to capture logical structure in statements. TAPAS~\cite{DBLP:conf/acl/HerzigNMPE20}, and DATER~\cite{ye2023large} encode sentence-table pairs and transform table-based reasoning into question-answering or inference tasks.
Existing datasets primarily focus on specific domains like Wikipedia and finance. HybridQA~\cite{chen-etal-2020-hybridqa} derived from Wikipedia emphasises span lookup, while TAT-QA~\cite{zhu2021tatqa}, FinQA~\cite{chen-etal-2021-finqa}, and DocMath-Eval~\cite{zhao-etal-2024-docmath} address numerical reasoning in the financial domain. SciGen~\cite{moosavi2021scigen} introduces a scientific table-to-text generation dataset that requires arithmetic reasoning, but focuses mainly on generation rather than comprehensive reasoning evaluation. However, these datasets have relatively limited reasoning types, significantly differing from real-world scientific table understanding that require numerical computation reasoning. To address this gap, we propose MMSci dataset that combines multiple reasoning types to enhance model performance on complex scientific table understanding tasks.

% \begin{table}[h]
% \centering
% \resizebox{0.79\linewidth}{!}
% {\begin{tabular}{lr}
%     \toprule
%     \textbf{Type} & \textbf{Value}\\
%     % \midrule
%     % \multicolumn{2}{c}{\textbf{\textit{Basic Insight}}} \\
%      \midrule
%     Tables & 3681 \\
%     Question Length(Avg) & 20.30 \\
%     Answer Length (Avg) & 8.52 \\
%     % \textbf{Columns Per Table} & 6.68 \\
%     % \textbf{Rows Per Table } & 16.71 \\
%     % \textbf{Ratio of Numerical Cells} & 65.74\% \\
%     Average Reasoning Steps & 6.26 \\
%     Training Set & 886 \\
%     Testing Set & 19,661 \\
%     \bottomrule
% \end{tabular} 
% }
% \caption{
% The reasoning types, the description of their subtypes, and their proportion in our dataset. 
% The number in parentheses is the proportion of each reasoning type.
% }
% \label{table:dataset_statistic}
% % \vspace{-15pt}
% \end{table}

\section{Construction of MMSci Dataset}
As shown in Figure~\ref{fig:model}, our framework consists of four key stages: (1) Table Image Generation: converting SciGen dataset tables into high-quality HTML and image formats; (2) MMSci Dataset Construction: employing GPT-4o~\cite{OpenAI2024} with self-consistency voting~\cite{wang_self-consistency_2023} and human verification to generate high-quality samples; (3) Table Structure Learning: training MLLMs to generate HTML representations from table images; and (4) Visual Instruction Tuning: fine-tuning MLLMs on instruction-following data across TQA, TFV, and T2T tasks.
\subsection{Data Collection}
To construct MMSci dataset, we focus on scientific tables that contain significant numerical values and complex reasoning requirements. We collect raw tabular data from the SciGen dataset~\cite{moosavi2021scigen}, which provides pairs of scientific tables and their corresponding descriptions across computer science research domains. A large number of these descriptions require arithmetic reasoning
% (e.g., argMax, argMin, comparison, subtraction) 
over table values, indicating the natural presence of numerical reasoning in scientific table descriptions.
We transform the original textual tables into high-quality HTML format and then render them into table images while preserving their structural integrity. This process ensures that the visual representation maintains the complex layouts and relationships present in the original scientific tables. Finally, we collect 52K image-to-HTML pairs based on tables from the training set and development set of the SciGen dataset.

\subsection{Table Structure Learning}
% Existing table-based MLLMs~\cite{lee_pix2struct_2023, alonso-etal-2024-pixt3, zheng_multimodal_2024} demonstrate that generating textual table representations given the table image is crucial for aligning table structure and text information within the table image. Therefore, we create 52K instruction-following image-to-HTML samples based on the tables from the training set of SciGen dataset via the Imgkit\footnote{\url{https://pypi.org/project/imgkit/}} python package. Each sample consists of a table image with their corresponding HTML format of table representations. The resulting pre-training data contains 52K table image-to-HTML examples, which is denoted as MMSci-pre dataset.

Existing table-based MLLMs~\cite{lee_pix2struct_2023, alonso-etal-2024-pixt3, zheng_multimodal_2024} demonstrate that generating textual table representations given the table image is crucial for aligning table structure and text information within the table image. Therefore, we create 52K instruction-following image-to-HTML samples based on the tables from the training set of SciGen dataset via the Imgkit\footnote{\url{https://pypi.org/project/imgkit/}} python package. The resulting pre-training data contains 52K table image-to-HTML examples, which is denoted as MMSci-Pre dataset.

\subsection{Numerical Reasoning Augmentation}

For the construction of instruction dataset MMSci-Ins and MMSci-Eval, we select 12,000 tables from the training set and 1,038 from the testing set of SciGen dataset to create MMSci-Ins and MMSci-Eval datasets, respectively. For each table, we employ GPT-4o to generate task-specific content by feeding the table image and its corresponding descriptions in SciGen dataset. For TQA tasks, we generate questions paired with corresponding reasoning steps and answers. For TFV tasks, we create claims along with supporting reasoning steps and verification results. The labels of claim including three types: supported, refuted, and not enough information, which is consistent with the existing TFV datasets~\cite{lu-etal-2023-scitab}. Additionally, we augment the existing table-to-text pairs in SciGen dataset with detailed reasoning steps for T2T tasks.

To ensure the quality and consistency of the generated content, we employ the self-consistency Chain-of-Thought (CoT) reasoning mechanism~\cite{wang_self-consistency_2023}. For each task, we generate multiple reasoning paths and employ a voting mechanism to determine the final output. To validate the quality of these generated samples, we employ a two-stage verification process: first using GPT-4o to assess the consistency between reasoning steps and outputs, then manually assess 40\% of the generated samples to further ensure high quality. The identified false samples are regenerated using GPT-4o to maintain dataset quality. 
% For the instruction dataset construction, we select 1,2000 table images and their corresponding descriptions from the SciGen dataset. For each table, we employ GPT-4o~\cite{OpenAI2024} to generate task-specific content by feeding the table image and its corresponding descriptions in SciGen dataset. In TQA tasks, we generate questions paired with corresponding reasoning steps and answers. For TFV tasks, we create claims along with supporting reasoning steps and verification results. The labels of claim including three types: (supported, refuted, and not enough information, which is consistent with the exisiting TFV datasets~\cite{lu-etal-2023-scitab}. Additionally, we augment the existing table-to-text pairs in SciGen with detailed reasoning steps for T2T tasks. 
% To ensure the quality and consistency of the generated content, we implement a self-consistency Chain-of-Thought (CoT) reasoning mechanism~\cite{wang2022self}. For each task, we generate multiple reasoning paths and employ a voting mechanism to determine the final output. This approach helps mitigate potential errors and ensures robust reasoning processes.To validate the quality of these generated samples, we employ a two-stage verification process. First, we utilise GPT-4o to assess the consistency between the reasoning steps and the final answers or claims. Subsequently, we manually annotate 30\% of the generated samples to ensure high quality, with particular attention to the correctness of reasoning steps and final outputs.
Through this rigorous construction and verification process, we create the MMSci-Ins dataset comprising 12K instruction-tuning samples based on the training set of SciGen dataset, and the MMSci-Eval benchmark with 3,114 testing examples based on the testing set of SciGen dataset. Each sample includes detailed step-by-step reasoning processes, enabling models to learn both the final outputs and the logical progression needed to arrive at those conclusions. The dataset maintains a balanced distribution across the three tasks, where each table is paired with one TQA sample, one TFV sample, and one T2T sample, ensuring comprehensive coverage of different reasoning requirements in scientific table understanding. Detailed dataset quality control could be found in the Appendix~\ref{app:quality}.
% Through this rigorous data construction process, we create the MMSci-Ins dataset comprising 12K instruction-tuning samples based on the training set of SciGen dataset, and the MMSci-Eval benchmark with 3,114 testing examples based on the testing set of SciGen dataset. Each sample includes detailed step-by-step reasoning processes, enabling models to learn both the final outputs and the logical progression. The dataset maintains a balanced distribution across the three tasks where each table is paired with one TQA, one TFV and one T2T task, ensuring comprehensive coverage of different reasoning requirements in scientific table understanding.

%--------------------------------dataset--------
%--------------------------------dataset--------
% \input{tables/dataset}
%--------------------------------dataset--------
%--------------------------------dataset--------

%--------------------------------MMSci--------
%--------------------------------MMSci--------

\begin{table*}[ht]\footnotesize
% \small
\centering
\renewcommand{\arraystretch}{1.1}
\setlength\tabcolsep{20pt}
\resizebox{0.83\linewidth}{!}{
\begin{tabular}{lccccc} 
\toprule
\multirow{3}{*}{\textbf{Models}} 
& \multicolumn{3}{c}{\textbf{MMSci-Eval}}&\multicolumn{2}{c}{\textbf{Held-out}}\\  
% \cline{2-6}
 & \textbf{TQA} & \textbf{TFV} & \textbf{T2T} & \multicolumn{1}{c}{\textbf{TABMWP}} & \textbf{TAT-QA} \\ 
 % \cline{2-6}
 &\multicolumn{1}{c}{Acc.} & \multicolumn{1}{c}{Acc.} & \multicolumn{1}{c}{BLEU} & \multicolumn{1}{c}{Acc.} & \multicolumn{1}{c}{Acc.} \\
\hline
% \multicolumn{6}{l}{{\cellcolor[rgb]{0.957,0.957,0.957}}MMSci-Ins-3000} \\
\multicolumn{6}{l}{{\cellcolor[rgb]{0.957,0.957,0.957}}\textbf{Baseline}} \\
\multicolumn{1}{l}{GPT-4V~\citep{OpenAI2023}} & \textbf{53.13} & \textbf{78.01} & \textbf{4.80} & \textbf{60.00} & \textbf{32.50} \\\hline\hline
\multicolumn{1}{l}{InternVL-2-76B~\citep{chen2024internvl}} & 40.31 & 62.46 & 1.79 &  46.28 & 6.73 \\
\multicolumn{1}{l}{LLaVA-NeXT-72B~\citep{li2024llavanext-strong}} & 11.75 & 49.28 & 1.79 & 10.69 & 3.29 \\
\multicolumn{1}{l}{Qwen-2-VL-72B-Ins.~\citep{wang2024qwen2}} & 39.11 & 64.06 & 2.83 & 41.42 & 17.65 \\
\hline
\multicolumn{1}{l}{LLaVA-NeXT-34B~\citep{li2024llavanext-strong}} & 9.73 & 42.19 & 2.33 & 6.96 & 1.29 \\
\multicolumn{1}{l}{LLaVA-NeXT-13B~\citep{li2024llavanext-strong}} & 2.31 & 1.83 & 1.79 & 1.67 & 0.43 \\
\multicolumn{1}{l}{Table-LLaVA-13B~\cite{zheng_multimodal_2024}} & 8.57 & 51.15 & 0.03 & \underline{59.77} & 15.67 \\
\multicolumn{1}{l}{Pixtral-12B~\cite{agrawal2024pixtral}} & 0.96 & 5.49 & 4.12 & 4.64 & 7.46 \\
\multicolumn{1}{l}{Llama-3.2-11B-Vision-Ins.~\cite{meta2024llama3.2}} & 1.15 & 5.85 & 3.04 & 7.39 & 0.37 \\
\hline
\multicolumn{1}{l}{LLaVA-NeXT-7B~\cite{li2024llavanext-strong}} & 0.19 & 0.86 & 2.99 & 1.73 & 0.72 \\
\multicolumn{1}{l}{Qwen-2-VL-7B-Ins.~\citep{wang2024qwen2}} & 25.62 & 52.79 & 3.04 &34.43 & 16.19\\
\multicolumn{1}{l}{InternVL-2-8B~\citep{chen2024internvl}} & 25.72 & 44.99 & 2.64 &  18.42 &7.12 \\
\multicolumn{1}{l}{MiniCPM-V-2.6-8B~\cite{yao2024minicpm}} & 26.58 & 33.23 & 0.07 & 24.30 & 11.94 \\
\multicolumn{1}{l}{Table-LLaVA-7B~\cite{zheng_multimodal_2024}} & 7.99 & 39.30 & 0.03 & 57.78 & 12.82 \\
% \multicolumn{1}{l}{Table-LLaVA-13B} & 8.57 & 51.15 & 0.03 & 59.77 & 15.67 \\
\hline
%LLaVA-NEXT-7B
\multicolumn{6}{l}{{\cellcolor[rgb]{0.957,0.957,0.957}}\textbf{Ours  (LLaVA-NeXT-7B)}} \\
\multicolumn{1}{l}{MMSci-Pre (52k) + MMSci-Ins} & 17.72 & 57.12 & 2.93  & 49.47 & 10.46 \\
\multicolumn{1}{l}{MMTab-Pre (150k) + MMSci-Ins}  & 15.79 & 56.16 & 2.88 & 47.55 & 8.03 \\
% \multicolumn{1}{l}{MMTab-Pre+MMSci-Pre+MMSci-Ins} & 23.02 & 58.57 & 
\multicolumn{1}{l}{MM-Pre (202k) + MMSci-Ins} & 23.02 & 58.57 & 
2.36 & 49.72 & 12.27 \\
\multicolumn{1}{l}{w/o MM-Pre  (202k)}& 15.22 & 51.73 & 2.86 & 46.24& 7.63 \\
\hline\hline
%qwen7b
\multicolumn{6}{l}{{\cellcolor[rgb]{0.957,0.957,0.957}}\textbf{Ours (Qwen2-VL-7B-Ins.)}} \\
\multicolumn{1}{l}{MMSci-Pre (52k) + MMSci-Ins} & 41.13 & 72.92 & 3.24 & 49.50 & 19.68 \\
\multicolumn{1}{l}{MMTab-Pre (150k) + MMSci-Ins} & 40.75 & 72.73 & 3.16 & 49.08 & 19.30 \\
% \multicolumn{1}{l}{MMTab-Pre+MMSci-Pre (202k) +MMSci-Ins} & 
\multicolumn{1}{l}{MM-Pre  (202k) + MMSci-Ins} & 
\underline{42.10} & \underline{73.98} &\underline{3.29} & 49.96 & \underline{20.85}  \\
% \multicolumn{1}{l}{mixed pretrain  (202k) + MMTab-Ins} & 37.57 & 58.88 & 9.18 & 45.30 & 36.92 \\
% \multicolumn{1}{l}{w/o MMTab-Pre+MMSci-Pre (202k)} & 41.71 & 70.90 & 
\multicolumn{1}{l}{w/o MM-Pre  (202k)} & 41.71 & 70.90 & 
3.29 &48.02 &20.07 \\
\bottomrule
\end{tabular}
}

% \caption{We report average performance over benchmarks requiring numerical reasoning. `MMTab-Pre' stands for image-HTML pairs from . `successively IFT' represents that `LLaVA-Instruct' and `MMTab-Instruct' are used to fine-tune the model in a sequential order rather than mixed together. }
\caption{Performance comparison on MMSci-Eval and held-out tabular numerical reasoning datasets. MM-Pre (202k) indicates the combination of MMTab-Pre (150k) and MMSci-Pre (52k). w/o MM-Pre represents only training with MMSci-Ins dataset. Best results are in \textbf{bold}, second best are \underline{underlined}.}
\label{mmsci}

\end{table*}
%--------------------------------MMSci--------
%--------------------------------MMSci--------

%--------------------------------MMTab--------
%--------------------------------MMTab--------

\begin{table*}[t]
\small
% \footnotesize
% \centering
\renewcommand{\arraystretch}{1.5}
% \setlength\tabcolsep{5pt}
% \scalebox{0.89\linewidth}{
\resizebox{0.99\linewidth}{!}{

\begin{tabular}{lccccccccccccc} 
\toprule
\multicolumn{1}{c}{\multirow{3}{*}{\textbf{Method}}}
% % & \multirow{3}{*}{\textbf{LLM}} & \multicolumn{1}{c|}{\multirow{3}{*}{\textbf{Res.}}}
& \multicolumn{6}{c}{\textbf{TQA}} & \multicolumn{3}{c}{\textbf{TFV}} & \multicolumn{4}{c}{\textbf{T2T}} \\ 

% \cline{2-14}
 % & \multicolumn{1}{c}{\textbf{TABMWP}} & \multicolumn{1}{|c|}{\textbf{WTQ}} & \multicolumn{1}{c}{\textbf{HiTab}} & \multicolumn{1}{|c|}{\textbf{TAT-QA}} & \multicolumn{1}{c}{\textbf{FeTaQA}} & \multicolumn{1}{|c|}{\textbf{Avg. TQA}} & \multicolumn{1}{c}{\textbf{TabFact}} & \multicolumn{1}{|c|}{\textbf{InfoTabs}} &
 % \multicolumn{1}{c}{\textbf{Avg. TFV}} &\multicolumn{1}{|c|}{\textbf{HiTab\_T2T}} & \multicolumn{1}{c}{\textbf{Rotowire}} & \multicolumn{1}{|c|}{\textbf{WikiBIO}} & \multicolumn{1}{c}{\textbf{Avg. T2T}}\\ 
 & \multicolumn{1}{c}{\textbf{TABMWP}} & \multicolumn{1}{c}{\textbf{WTQ}} & \multicolumn{1}{c}{\textbf{HiTab}} & \multicolumn{1}{c}{\textbf{TAT-QA}} & \multicolumn{1}{c}{\textbf{FeTaQA}} & \multicolumn{1}{c}{\textbf{Avg. TQA}} & \multicolumn{1}{c}{\textbf{TabFact}} & \multicolumn{1}{c}{\textbf{InfoTabs}} &
 \multicolumn{1}{c}{\textbf{Avg. TFV}} &\multicolumn{1}{c}{\textbf{HiTab\_T2T}} & \multicolumn{1}{c}{\textbf{Rotowire}} & \multicolumn{1}{c}{\textbf{WikiBIO}} & \multicolumn{1}{c}{\textbf{Avg. T2T}}\\ 
% \cline{2-14}
 % &\multicolumn{1}{c}{Acc.} & \multicolumn{1}{|c|}{Acc.} & \multicolumn{1}{c}{Acc.} & \multicolumn{1}{|c|}{Acc.} & \multicolumn{1}{c}{BLEU} & \multicolumn{1}{|c|}{Acc.}
 % & \multicolumn{1}{c}{Acc.} & \multicolumn{1}{|c|}{Acc.} & \multicolumn{1}{c}{Acc.} & \multicolumn{1}{|c|}{BLEU} & \multicolumn{1}{c}{BLEU} & \multicolumn{1}{|c|}{BLEU} & \multicolumn{1}{c}{BLEU}  \\ 
  &\multicolumn{1}{c}{Acc.} & \multicolumn{1}{c}{Acc.} & \multicolumn{1}{c}{Acc.} & \multicolumn{1}{c}{Acc.} & \multicolumn{1}{c}{BLEU} & \multicolumn{1}{c}{Acc.}
 & \multicolumn{1}{c}{Acc.} & \multicolumn{1}{c}{Acc.} & \multicolumn{1}{c}{Acc.} & \multicolumn{1}{c}{BLEU} & 
 \multicolumn{1}{c}{BLEU} & \multicolumn{1}{c}{BLEU} & 
 \multicolumn{1}{c}{BLEU}  \\ 
\hline
\multicolumn{14}{l}{{\cellcolor[rgb]{0.957,0.957,0.957}}\textbf{Baseline}} \\
% \hline
\multicolumn{1}{l}{GPT-4V~\citep{OpenAI2023}} & \textbf{60.50} & \textbf{48.00} & \textbf{27.50} &\textbf{32.50} &11.04 & \textbf{35.91} &
45.50 &\underline{65.60} & 55.55 
&2.98 &4.23 &1.94&3.05 \\
\hline\hline
\multicolumn{1}{l}{Qwen2-VL-7B-Ins.~\citep{wang2024qwen2}} &34.44& 12.55& 3.36& 16.19& 11.75 &  15.66
&20.28& 34.19 &  27.23
&1.90& 2.30& 2.94 &  2.38\\
\multicolumn{1}{l}{LLaVA-NeXT-7B~\cite{li2024llavanext-strong}} &1.73& 0.00& 0.00& 0.00& 1.17 &  0.58
&1.24& 1.78 &  1.51
&0.45& 1.04& 0.67 &  0.72 \\
\multicolumn{1}{l}{Table-LLaVA-7B~\cite{zheng_multimodal_2024}}  & 57.78 & 18.43 & 10.09 & 12.82 & \underline{25.60} & 24.94
& \underline{59.85} & 65.26 & \underline{62.56}
& \underline{9.74} & \textbf{10.46} & \textbf{9.68} & \textbf{9.96} \\
\multicolumn{1}{l}{Table-LLaVA-13B~\cite{zheng_multimodal_2024}} & \underline{59.77} & \underline{20.41} & \underline{10.85} & 15.67 & \textbf{28.03} & \underline{26.95} 
& \textbf{65.00 }& \textbf{66.91} & \textbf{65.96}
& \textbf{10.40} & \underline{8.83} & \underline{9.67} & \underline{9.63}\\
\hline
%gray line

% \hline
%LLaVA-NEXT-7B
\multicolumn{14}{l}{{\cellcolor[rgb]{0.957,0.957,0.957}} \textbf{Ours (LLaVA-NeXT-7B)}} \\
% \multicolumn{1}{l}{direct inference} &1.05 & 1.50 & 0.72&2.90&0.12\\
\multicolumn{1}{l}{MMSci-Pre (52k) + MMSci-Ins} &8.76& 3.22& 0.63& 0.39& 5.99 &  3.80
&35.78& 25.37 &  30.57
&1.57& 1.10& 1.78 &  1.48
\\
\multicolumn{1}{l}{MMTab-Pre (150k) + MMSci-Ins} &9.00& 2.62& 0.63& 0.26& 7.23 &  3.95
&36.22& 26.91 &  31.56
&1.64& 0.84& 1.57 &  1.35
\\
% \multicolumn{1}{l}{MMTab-Pre+MMSci-Pre}+ MMSci-Ins &10.66& 4.83& 
\multicolumn{1}{l}{MM-Pre (202k) + MMSci-Ins} &10.66& 4.83& 
0.82& 0.65& 9.39 &  5.27
&39.63& 27.63 &  33.63
&1.13& 0.83& 1.90 &  1.29
\\
% \multicolumn{1}{l}{mixed pretrain + MMTab-Ins} & 28.35 & 43.06 & 8.48 & \underline{47.65} & \underline{38.01} \\
\multicolumn{1}{l}{w/o MM-Pre (202k)} &9.69& 2.74& 0.19& 0.39& 6.84 &  3.97
&31.72& 23.80 &  27.76
&1.69& 0.79& 1.53 &  1.34
\\
\hline\hline
%qwen2vl7b
\multicolumn{14}{l}{{\cellcolor[rgb]{0.957,0.957,0.957}}\textbf{Ours (Qwen2-VL-7B-Ins.)}} \\
\multicolumn{1}{l}{MMSci-Pre (52k) + MMSci-Ins} & 49.51&18.74&4.95&19.69&12.89 & 21.15
&37.93& 45.33 &  41.63
&0.75& 2.81& 2.69 &  2.08 \\
\multicolumn{1}{l}{MMTab-Pre (150k) + MMSci-Ins}&49.09& 18.95& 4.63& 19.30& 9.77 &  20.35
&40.00& 46.56 &  43.28
&0.91& 1.26& 2.89 &  1.69 \\
\multicolumn{1}{l}{MM-Pre (202k) + MMSci-Ins} &46.97& 19.73& 4.38& \underline{20.85}& 12.34 &  20.85
&39.99& 45.96 &  42.97
&0.96& 1.32& 2.60 &  1.63\\
% \multicolumn{1}{l}{mixed pretrain (202k) + MMTab-Ins} & 37.57 & 58.88 & 9.18 & 45.30 & 36.92 \\
\multicolumn{1}{l}{w/o MM-Pre (202k)} &48.02& 18.67& 5.33& 20.08& 12.58 &  20.94
&33.53& 44.93 &  39.23
&0.71& 2.76& 2.70 &  2.06 \\
\bottomrule
\end{tabular}
}
\caption{Performance comparison on MMTab held-out datasets. Best results are in bold, second best are underlined.}
\label{MMTab}
\end{table*}
\section{Experiments}
\subsection{Model Training}
To demonstrate the effectiveness of MMSci dataset, we train two series of MLLM following the architecture of Qwen2-VL-7B-Instruct~\cite{wang2024qwen2} and LLaVA-NeXT-7B~\cite{li2024llavanext-strong}.

\noindent\textbf{Model Architectures.} Both models follow a three-component design:
\textbf{Qwen2-VL-7B-Instruct} consists of a Vision Transformer (ViT)~\cite{dosovitskiy2020image} as the vision tower, a MLP as the vision-language connector, and a Qwen2-7B-Instruct~\cite{yang2024qwen2technicalreport} as the language model. \textbf{LLaVA-NeXT-7B} uses a pre-trained CLIP model~\citep{clip_paper} as the visual encoder, a MLP connector, and a Vicuna-7B model~\citep{vicuna2023} as the backbone. In both architectures, the vision encoder processes images into visual features, which are projected into the LLM's word embedding space via the MLP connector.

% \noindent\textbf{Training Process.}
% Following Table-LLaVA~\cite{zheng_multimodal_2024}, 
We divide the training into two stages:

\noindent\textbf{Table Structure Learning.} We use both MMSci-Pre and MMTab-Pre~\cite{zheng_multimodal_2024} corpus (202K table image-to-HTML pairs in total) to align visual features with textual representations in different experimental settings as shown in Table~\ref{mmsci}. Models learn to generate HTML table representations, developing table structure perception capabilities. For LLaVA-NeXT-7B, only the MLP connector parameters are updated during this stage.

\noindent\textbf{Visual Instruction Tuning.} We use 12K instruction-following samples from MMSci-Ins to fine-tune the MLLMs while keeping visual encoders frozen. Only the MLP projection layer and LLM weights are updated, focusing on developing instruction-following numerical reasoning capabilities across TQA, TFV, and T2T tasks.

Notably, both models support dynamic input resolutions, addressing a key limitation of existing table MLLMs~\cite{lee_pix2struct_2023, alonso-etal-2024-pixt3, zheng_multimodal_2024} that require fixed-size input image resolutions (e.g., 336×336). Qwen2-VL achieves this through 2D-RoPE~\cite{su2024roformer} to capture two-dimensional positional information of images while LLaVA-NeXT employs a simpler approach of splitting images into grids and encoding them independently.

\subsection{Experimental Settings}
\noindent\textbf{Baselines.}~~We select several state-of-the-art MLLMs as our baselines, including GPT-4V~\cite{OpenAI2023}, InternVL-2-76B~\cite{chen2024internvl}, LLaVA-NeXT series (72B/34B/13B/7B)~\cite{li2024llavanext-strong}, Qwen-2-VL-Instruct series (72B/7B)~\cite{wang2024qwen2}, Table-LLaVA series (13B/7B)~\cite{zheng_multimodal_2024}, Pixtral-12B~\cite{agrawal2024pixtral}, Llama-3.2-11B-Vision-Ins.~\cite{meta2024llama3.2}, MiniCPM-V-2.6-8B~\cite{yao2024minicpm}, and InternVL-2-8B~\cite{chen2024internvl}.
% These models represent diverse architectures and parameter scales, from 7B to 76B parameters, enabling comprehensive evaluation of our approach across different model capacities and design choices. GPT-4V serves as our primary baseline due to its strong performance on multimodal tasks, while open-source models like LLaVA-NeXT and Qwen-2-VL-Instruct provide important comparisons for reproducible research.

\noindent\textbf{Datasets and Metrics.}~~ The held-in evaluation sets in Table~\ref{mmsci} include TQA, TFV and T2T tasks of MMSci-Eval. The held-out evaluation sets in Table~\ref{MMTab} are from MMTab-Eval benchmark~\cite{zheng_multimodal_2024}. TQA contains TABMWP~\cite{tabmwp}, WTQ~\cite{WTQ}, HiTab~\cite{HiTab}, TAT-QA~\cite{zhu2021tatqa}, and FeTaQA~\cite{FeTaQA}, where TABMWP and TAT-QA specifically focus on tabular numerical reasoning. TFV contains TabFact~\cite{TabFact} and InfoTabs~\cite{infotabs}, while Table-to-Text (T2T) generation uses HiTab\_T2T~\cite{HiTab}, Rotowire~\cite{rotowire}, and WikiBIO~\cite{wikibio}. While these datasets contain tables from Wikipedia, financial reports, and government documents, our MMSci-Eval datasets primarily feature scientific tables with numerical values from research papers.
We use accuracy and BLEU~\cite{Papineni2002BleuAM} for TQA, TFV, and T2T benchmarks. 

\section{Results and Analysis}

\subsection{Performance on Numerical Reasoning Datasets}
The experimental results demonstrate the effectiveness of our proposed approach across various multimodal table understanding tasks. As shown in Table~\ref{mmsci}, we compare our method with state-of-the-art baselines on both MMSci benchmarks (TQA, TFV, T2T) and held-out tabular numerical reasoning datasets (TABMWP, TAT-QA).
Among the baseline models, GPT-4V~\cite{OpenAI2023} achieves superior performance across all tasks, establishing strong benchmarks with 53.13\% accuracy on TQA, 78.01\% on TFV, and notably strong generalisation ability on held-out numerical reasoning datasets. Large-scale open-sourced models like InternVL-2-76B~\cite{chen2024internvl} and Qwen-2-VL-72B~\cite{wang2024qwen2} also demonstrate competitive performance but show relatively weaker generalisation to held-out numerical reasoning datasets.

As for our approaches, with LLaVA-NeXT-7B as the foundation model, we observe that training with MMSci-Pre (52k) dataset demonstrates higher performance (17.72\% on TQA, 57.12\% on TFV) compared to training with MMTab-Pre (150k) dataset (15.79\% on TQA, 56.16\% on TFV). The combination of both table structure learning dataset (MM-Pre 202k) further improves performance to 23.02\% on TQA and 58.57\% on TFV. Notably, our approach shows strong generalisation ability on held-out datasets, achieving 49.72\% on TABMWP with the experiment setting of MM-Pre (202k) + MMSci-Ins.

With Qwen2-VL-7B-Instruct as the foundation model, we observe significantly stronger performance across all settings. Training with MMSci-Pre (52k) + MMSci-Ins achieves comparable or better performance (41.13\% on TQA, 72.92\% on TFV) compared to training with MMTab-Pre (150k) + MMSci-Ins (40.75\% on TQA, 72.73\% on TFV), despite using only one-third of the table structure learning data. The experiment setting of training with MM-Pre (202k) + MMSci-Ins achieves the best performance with 42.10\% accuracy on TQA and 73.98\% on TFV, while also demonstrating strong generalisation ability on held-out numerical reasoning datasets (49.96\% on TABMWP and 20.85\% on TAT-QA).

%%%%%--------------\input{tables/abl}

\begin{table}[t]\footnotesize
\small
\centering
\resizebox{0.95\linewidth}{!}{
\begin{tabular}{lccccc}
\toprule
\multirow{2.5}{*}{\textbf{Models}} & \multicolumn{3}{c}{\textbf{MMSci-Eval}}&\multicolumn{2}{c}{\textbf{Held-out}}\\  
% \textbf{Model} & 
&\textbf{TQA} & \textbf{TFV} & \textbf{T2T} & \textbf{TABMWP} & \textbf{TAT-QA} \\ 
\midrule
% \textbf{Qwen2-VL}                & 97.80        & 16.38       & -           & 48.09       & 16.18       \\
% \rowcolor{grey!25}\textbf{w \textsc{MCDGraph}}              & 98.34      & 25.92    & -           & 60.94      & 25.44       \\
% \textbf{InternVL2}               & 77.45        & 9.78        & -           & 50.75       & 25.01        \\

%LLaVA-NEXT-7B
\multicolumn{6}{l}{{\cellcolor[rgb]{0.957,0.957,0.957}}\textbf{Ours (LLaVA-NeXT-7B)}} \\
\rowcolor{green!8}{MMSci-Pre (52k) + MMSci-Ins} & 17.72 & 57.12 & 2.93  & 49.47 & 10.46 \\
\rowcolor{blue!8}{w/o Reasoning} & 10.75 & 42.73 & 2.16 & 42.50 & 7.68 \\
\hline

\rowcolor{green!8}{MMTab-Pre (150k) + MMSci-Ins} & 15.79 & 56.16 & 2.88 & 43.55 & 8.03 \\
\rowcolor{blue!8}{w/o Reasoning} & 9.58 & 50.31 & 1.93 & 42.50 & 7.42 \\
\hline

% \rowcolor{green!8}{MMTab-Pre+MMSci-Pre+MMSci-Ins} & 23.02 & 58.57 & 
\rowcolor{green!8}{MM-Pre (202k) + MMSci-Ins} & 23.02 & 58.57 & 
2.36 & 49.72 & 12.27 \\
\rowcolor{blue!8}{w/o Reasoning} & 12.73 & 45.21  & 2.16 & 46.50 & 19.68 \\
\hline

% \rowcolor{green!8}{w/o MMTab-Pre+MMSci-Pre}  & 15.22 & 51.73 & 2.86 & 
\rowcolor{green!8}{w/o MM-Pre (202k)}  & 15.22 & 51.73 & 2.86 & 
46.24& 7.63 \\
\rowcolor{blue!8}{w/o Reasoning} & 9.43 & 42.31 & 2.36 & 45.50 & 8.39 \\
\hline\hline
% \hline
\multicolumn{6}{l}{{\cellcolor[rgb]{0.957,0.957,0.957}}\textbf{Ours (Qwen2-VL-7B-Ins.)}} \\
\rowcolor{green!8}{MMSci-Pre (52k) + MMSci-Ins} & 41.13 & 72.92 & 3.24   & 49.50 & 19.68 \\
\rowcolor{blue!8}{w/o Reasoning} & 35.06 & 66.47 & 3.14 & 44.08 & 16.72 \\
\hline

\rowcolor{green!8}{MMTab-Pre (150k) + MMSci-Ins} & 40.75 & 72.73& 3.16 & 49.08 & 19.30 \\
\rowcolor{blue!8}{w/o Reasoning} & 34.48 & 66.28 & 2.27 & 43.97 & 16.07 \\
\hline

\rowcolor{green!8}{MM-Pre (202k) + MMSci-Ins} & \underline{42.10} & \underline{73.98} &3.29 & 49.96 & 20.85  \\
\rowcolor{blue!8}{w/o Reasoning} & 35.45 & 67.43 & 1.97 & 46.34 & 17.68 \\
\hline

% \rowcolor{green!8}{w/o MMTab-Pre+MMSci-Pre(202k)} & 41.71 & 70.90 & 
\rowcolor{green!8}{w/o MM-Pre (202k)} & 41.71 & 70.90 & 
3.29 &48.02 &20.07 \\
\rowcolor{blue!8}{w/o Reasoning} & 34.44 & 62.90 & 3.18 & 44.60 & 14.68 \\

\bottomrule
\end{tabular}}
% \vspace{-0.2cm}
\caption{Ablation study results for reasoning steps on MMSci-Eval and held-out datasets. 
% Results demonstrate the effectiveness of reasoning steps across model architectures, with Qwen2-VL-7B showing superior performance and strong generalization ability.
}
\label{tab:ablation}
\end{table}
%%%%%--------------\input{tables/abl}

These results demonstrate that our proposed MMSci-Pre dataset with 52K scientific domain-specific data is more effective than MMTab-Pre with 150K general-domain data, highlighting the importance of data quality over quantity. Furthermore, Qwen2-VL-7B-Instruct consistently outperforms LLaVA-NeXT-7B across all experimental settings, suggesting its stronger capability in table understanding and numerical reasoning tasks. Besides, our approach shows strong generalisation to held-out tabular numerical reasoning datasets, demonstrating enhanced general ability in multimodal table understanding and reasoning.

\begin{figure*}[htb]
% \small
\centering
\includegraphics[width=0.83\linewidth]{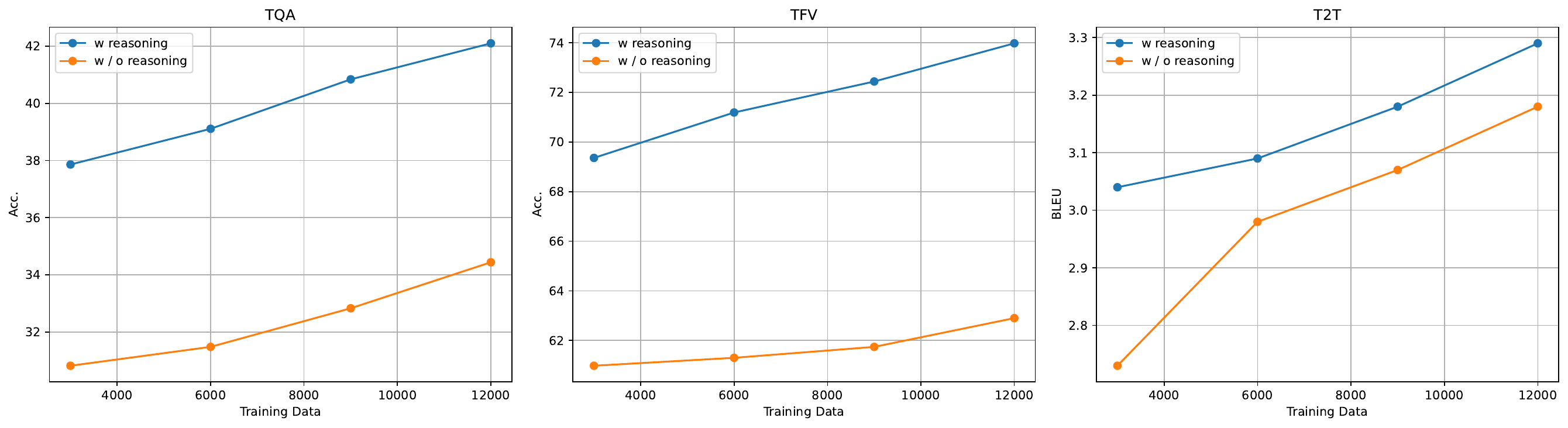}
\caption{Performance scaling with increasing instruction tuning data size on three MMSci tasks. 
}
\label{fig:scaling}
\end{figure*}

\subsection{Performance on Held-out MMTab Benchmarks}

The experimental results in Table~\ref{MMTab} also demonstrate the effectiveness and generalisation ability of our proposed approach across various held-out MMTab benchmark. 
% table understanding tasks. As shown in Table~\ref{MMTab}, we evaluate our method against state-of-the-art baselines on MMTab held-out datasets across three primary tasks: TQA, TFV, T2T.
GPT-4V~\cite{OpenAI2023} show strong performance across all tasks, achieving 35.91\% average accuracy on TQA, 55.55\% on TFV, and 3.05 BLEU on T2T. Table-LLaVA models, which are specifically trained on MMTab-Ins dataset, demonstrate competitive performance. Table-LLaVA-13B achieves strong results on TFV (65.96\% average accuracy) and T2T (9.63 BLEU) while Table-LLaVA-7B shows robust performance on TABMWP (57.78\%).

As for our approaches, with LLaVA-NeXT-7B as the foundation model, we observe that training with MMSci-Pre (52k) and MMSci-Ins, despite not being trained on MMTab-Ins dataset~\cite{zheng_multimodal_2024}, demonstrates promising generalisation ability. The MMSci-Pre (52k) + MMSci-Ins combination achieves 3.80\% average accuracy on TQA and 30.57\% on TFV with only scientific domain data. The combination of both table structure learning datasets (MM-Pre 202k) further improves performance across all metrics, reaching 5.27\% on TQA and 33.63\% on TFV.
As for Qwen2-VL-7B-Instruct as the foundation model, we observe significantly stronger generalisation capability. MMSci-Pre (52k) + MMSci-Ins combination achieves 21.15\% average accuracy on TQA and 41.63\% on TFV, demonstrating strong zero-shot transfer to MMTab benchmark despite using only scientific domain data (MMSci dataset). This performance is particularly impressive when compared to MMTab-Pre (150k) + MMSci-Ins combination, which uses three times more image-to-HTML data. Even without any table structure learning (w/o MM-Pre), our approach achieves competitive results, highlighting the effectiveness of our MMSci-Ins instruction tuning dataset.

These results empirically demonstrate our MMSci-Pre dataset with 52K scientific domain-specific data achieves comparable or better performance than MMTab-Pre with 150K general-domain data in MMTab held-out benchmark, highlighting the importance of scientific domain-specific tables. Even without MMTab table structure learning data, our approach demonstrates strong generalisation ability, particularly evident in the performance of MMSci-Pre (52k) + MMSci-Ins and w/o MM-Pre experiment settings. 
% Finally, Qwen2-VL-7B-Ins. shows superior generalisation capability compared to LLaVA-NeXT-7B, suggesting its stronger potential for multimodal table understanding tasks.

%%%%%--------------\input{tables/unembed_align}
\begin{table*}[ht!]\footnotesize
%\scriptsize
\setlength\tabcolsep{8pt}
\centering
\resizebox{0.8\linewidth}{!}{
% \small
\renewcommand\arraystretch{1.1}
\begin{tabular}{lccccccc} %p{1.5cm}p{1.5cm}p{1.5cm}p{1.5cm}p{1.5cm}p{1.5cm}
\toprule
\textbf{Models} & \textbf{Cycle KNN} & \textbf{Mutual KNN} & \textbf{Lcs KNN} & \textbf{CKA} & \textbf{CKNNA} & \textbf{SVCCA} & \textbf{Edit KNN} \\ \hline
\multicolumn{8}{c}{\textbf{\textit{Unembedding stage: ImageNet(Concepts)}}} \\
% \hline
Random & 0.02761 & 0.01257 & 0.52355 & 0.08614 & 0.00714 & 0.12425 & 0.00019\\
% Qwen2-VL-2B-Instruct & 0.50947 & 0.02844 & 1.06713 & 0.05075 & 0.01500 & 0.10455 & 0.00047 \\
Qwen2-VL-7B-Ins. & \textcolor{black}{\textbf{0.68110}} & \underline{0.03486} & \underline{1.28153} & \underline{0.08856} & \textcolor{black}{\textbf{0.03067}} & \textcolor{black}{\textbf{0.14318}} & \textcolor{black}{\textbf{0.00112}} \\
Llama3.2-11B-Vision-Ins. & \underline{0.08608} & \textcolor{black}{\textbf{0.04205}} & \textcolor{black}{\textbf{1.52788}} & 0.06079 & 0.01403 & 0.11651 & 0.00061 \\
LLaVA-NeXT-7B & 0.57173 & 0.02077 & 0.81645 & 0.08024 & 0.01577 & 0.13240 & 0.00037 \\
Phi3.5-Vision-Ins. & 0.02761 & 0.01257 & 0.52355 & 0.08614 & 0.00714 & 0.12118 & 0.00019 \\
InternVL2-8B & 0.08175 & 0.01637 & 0.72495 & \textcolor{black}{\textbf{0.09185}} & 0.00062 & 0.12148 & 0.00044\\ \hline
\hline
\multicolumn{8}{c}{\textbf{\textit{Unembedding stage: Wikipedia Caption (short descriptive sentences)}}} \\
% \hline
% Qwen2-VL-2B-Instruct & 0.33496 & 0.04751 & 1.47167 & 0.02368 & 0.00529 & 0.17856 & 0.00048 \\
Qwen2-VL-7B-Ins. & \underline{0.49414} & \textcolor{black}{\textbf{0.06855}} & \textcolor{black}{\textbf{2.05078}} & \textcolor{black}{\textbf{0.08876}} & \textcolor{black}{\textbf{0.04093}} & 0.20229 & \textcolor{black}{\textbf{0.00175}} \\
Llama3.2-11B-Vision-Ins. & 0.31347 & 0.03623 & 1.29980 & 0.00968 & 0.00779 & \underline{0.22120} & 0.00050 \\
LLaVA-NeXT-7B & \textcolor{black}{\textbf{0.57813}} & 0.03935 & 1.36523 & \underline{0.07933} & \underline{0.03998} & \textcolor{black}{\textbf{0.23114}} & \underline{0.00082} \\
Phi3.5-Vision-Ins. & 0.04980 & 0.03027 & 1.14843 & 0.01669 & 0.03890 & 0.18183 & 0.00066 \\
InternVL2-8B & 0.36914 & 0.04132 & \underline{1.55761} & 0.04732 & 0.01658 & 0.21739 & 0.00093 \\ \hline \hline
\multicolumn{8}{c}{\textbf{\textit{Embedding stage: MMSci T2T tasks (table to text description).}}} \\
Qwen2-VL-7B-Ins. & \textcolor{black}{\textbf{0.38631}} & \textcolor{black}{\textbf{0.06726}} & \underline{2.03660} & \textcolor{black}{\textbf{0.19318}} & \underline{0.05514} & \textcolor{black}{\textbf{0.38461}} & \textcolor{black}{\textbf{0.00183}} \\
Llama3.2-11B-Vision-Ins. & 0.31310 & 0.02200 & 0.84007 & 1.73979e-8 & 0.03208 & 0.08180 & 0.00026 \\
LLaVA-NeXT-7B & 0.38246 & 0.04514 & 1.49325 & 0.15203 & \textcolor{black}{\textbf{0.06673}} & \underline{0.28857} & 0.00109 \\
Phi3.5-Vision-Ins. & \underline{0.38053} & \underline{0.06712} & \textcolor{black}{\textbf{2.12909}} & \underline{0.16121} & 0.03688 & 0.26982 & \underline{0.00127} \\
InternVL2-8B & 0.36512 & 0.04651 & 1.56647 & 0.04230 & 0.02675 & 0.11876 & 0.00096 \\ 
\toprule
\end{tabular}
}
\caption{Kernel alignment analysis. The representation for each sample is the averaged token embeddings. The best two values are shown in \textbf{\textcolor{black}{bold}} and \underline{underlined}.
} 

% we can observe that Qwen2-VL-7B-Instruct generally outperforms other baselines on both datasets, which is aligned with the finding of Table \ref{tab:main}.} 
\label{tab:eval_unembed_align}
\end{table*}
%%%%%--------------\input{tables/unembed_align}

% \subsection{Ablation Study on Reasoning Steps}
\subsection{Ablation Study on Reasoning Steps}
We evaluate the effectiveness of reasoning steps across different experiment configurations. As shown in Table~\ref{tab:ablation},
% The results reveal interesting patterns in both model architectures and data efficiency.
Qwen2-VL-7B-Instruct demonstrates superior performance across all configurations. Without reasoning steps, the model training with MMSci-Pre (52k) + MMSci-Ins achieves better results than that with MMTab-Pre (150k) + MMSci-Ins, highlighting the importance of domain-specific table structure learning over data quantity. Adding reasoning steps consistently improves performance across all metrics, with the model reaching its peak performance under the MM-Pre (202k) + + MMSci-Ins experiment configuration.
Similar trends are observed in LLaVA-NeXT-7B, though with lower overall performance. These patterns extend to held-out tabular numerical reasoning datasets, where both models show strong generalisation capabilities with reasoning steps, especially on numerical reasoning tasks like TABMWP and TAT-QA. The results demonstrate that a smaller amount of scientific domain-specific table structure learning data, combined with explicit reasoning steps, can be more effective than larger-scale general domain table structure learning.

\subsection{Impact of Training Data Size}
% We investigate how model performance scales with the size of instruction tuning data and analyze the impact of reasoning steps. Figure \ref{fig:scaling} shows the performance trends across three tasks (TQA, TFV, T2T) as we increase the training data from 3K to 12K samples.

% The results demonstrate two key findings. First, models trained with reasoning steps consistently outperform their counterparts across all data sizes, with particularly large gaps in TQA (7-8\% absolute improvement) and TFV (8-10\% improvement). Second, while both variants benefit from increased training data, models with reasoning steps show stronger scaling behavior, suggesting that explicit reasoning helps models better utilize additional training examples.

% Notably, the performance gap between models with and without reasoning steps remains substantial even at larger data sizes, indicating that reasoning steps provide fundamental improvements in model learning that cannot be simply compensated for by additional training data.

As shown in Figure \ref{fig:scaling}, we compare performance of MLLMs training with MM-Pre (202k) + MMSci-Ins across three MMSci tasks (TQA, TFV, T2T) with instruction tuning data (MMSci-Ins dataset) size increasing from 3K to 12K samples. The findings demonstrate consistent advantages of incorporating reasoning steps across all data scales.
Models trained with reasoning steps maintain substantial performance advantages across all tasks (7-8\% for TQA, 8-10\% for TFV, 0.3-0.4 BLEU for T2T). While both variants benefit from increased training data, models with reasoning steps show stronger scaling behavior, particularly in TQA and TFV tasks. The persistent performance gap across all data sizes suggests that reasoning steps provide fundamental improvements in model learning that cannot be simply achieved through increased training data alone.

%-------------------------------------
% yingji: unembed alignment analysis
\subsection{Representational Alignment Analysis}
% Previous studies have highlighted that the alignment between language and vision modalities significantly impacts the performance of MLLMs in visual understanding tasks \cite{}. 
In this section, we conduct an in-depth analysis to assess the language-vision alignment from the perspective of the representation space. This analysis aims to provide further insights into the observed variations in model performance, particularly in the context of scientific multimodal table understanding and reasoning tasks.

\paragraph{Preliminaries.} We formalise MLLMs within the framework of an \textit{unembedding-embedding} architecture. In this framework, the unembedding stage is responsible for learning transformations between observations (e.g., text, vision) and latent spaces through encoders, while the embedding stage captures the complex interactions among latent variables within the latent space of LLMs' hidden layers. Each stage serves distinct functions and yields representations with different properties \citep{pmlr-v235-park24c}. Consequently, by focusing on each stage independently, we can have a systematical evaluation of model behaviours in representation spaces. To assess the representational alignment between vision-language modalities at each stage, we next measure the geometrical similarity between them via the \textit{kernel}. 

Kernels, characterising the distance metrics between points in a representation space, are commonly used to assess vector space \citep{pmlr-v235-huh24a}. Typically, the more similarity between two kernels derived from different spaces (text or vision) indicates a higher degree of alignment between those modality spaces. This similarity can be quantified via \textit{kernel-alignment metrics}, such as Centered Kernel Distance (CKA) \citep{kornblith2019similarity}. For more information about kernel-alignment metrics used in the experiment, we refer to \citet{pmlr-v235-huh24a} for a deep understanding. 

\paragraph{Quantitative evaluation.} For the unembedding stage, we specifically choose two language-vision datasets: ImageNet \citep{5206848} and Wikipedia Caption (WIT) \citep{srinivasan2021wit}. We randomly select 2048 samples from each dataset. These datasets offer varying levels of fine granularity in language-vision alignment, enabling a comprehensive assessment of representational performance. As illustrated in Table \ref{tab:eval_unembed_align}, we can observe that the Qwen2-VL-7B-Instruct  can generally outperform other baselines on both datasets, indicating it has better fine-grained alignment between language and vision. In the embedding stage, we evaluate alignment on the MMSci T2T task. Since some models do not support single-modality input, we utilise a reference language model (e.g., open-llama-
% 7B\footnote{\url{https://huggingface.co/openlm-research/open_llama_7b}})
7B~\cite{openlm2023openllama})
as the text encoder and MLLMs as the image encoder with prompt \textit{``please describe the table''}. Alignment is measured based on the output embedding from the last hidden layer. As shown in Table \ref{tab:eval_unembed_align}, Qwen2-VL-7B-Instruct outperforms the other models, demonstrating its superior language-vision alignment capability.
This segment of the experiment demonstrates that the Qwen2-VL-7B-Instruct model exhibits superior language-vision alignment within the representation space. This finding is consistent with the cross-modal consistency analysis presented in Appendix \ref{sec:consist}, where we evaluate different table information modalities as inputs to MLLMs and assess their cross-modal consistency (i.e., the proportion of identical predictions) on TQA and TFV tasks.

% yingji: embed (last hidden layer) alignment analysis
% Next, we focus on the embedding stage. We evaluate the kernel alignment of the last hidden output embeddings on our Table-Description dataset.

% \subsection{Case Study}
% To validate the effectiveness of AMR graphs in discriminating adversarial negative responses, we showcase several adversarial negative context-response pairs in Table~\ref{tab:case study1}. These samples are classified as ``positive'' by without the AMR graph information while can be correctly classified as ``negative'' when incorporating with the AMR graph information.

\section{Conclusion}
In this paper, we introduce a comprehensive framework for multimodal scientific table understanding and reasoning with dynamic input image resolutions.
% , consisting of MMSci-Pre, MMSci-Ins, and MMSci-Eval components.
% Our experiments demonstrate that domain-specific table structure learning with 52K scientific table images outperforms 150K general-domain tables, highlighting the importance of data quality over quantity in table understanding tasks.
Experimental results validate our framework's effectiveness across different model architectures, showing consistent improvements in both general table understanding and numerical reasoning capabilities, with strong generalisation to held-out datasets.

\section*{Limitations}
While this work advances scientific multimodal table understanding and reasoning, several limitations remain for future research. First, our framework primarily focuses on scientific tables containing numerical values, while other types of scientific tables (e.g., qualitative comparison tables, methodology tables) are not extensively covered. Second, though our framework demonstrates strong performance on numerical reasoning tasks, the current approach may still struggle with complex statistical analyses and domain-specific mathematical notations that are common in scientific literature. Third, while our models support dynamic input resolutions, processing extremely large tables with dense information remains challenging due to computational constraints and potential information loss during visual encoding.

\section*{Ethical Statement}
The MMSci datasets are constructed from publicly available scientific papers and their associated tables, primarily sourced from open-access repositories and academic databases with appropriate licenses. All table images are generated through automated scripts from the original scientific papers, maintaining their integrity while ensuring proper attribution. The instruction tuning samples are created based on the original scientific context, preserving the academic nature of the source material.
Our framework is designed to assist in scientific research by improving the accessibility and understanding of tabular data in academic literature. We anticipate that this work will contribute positively to the research community by facilitating more efficient analysis of scientific publications. The code and datasets are made publicly available for research purposes, promoting transparency and reproducibility in the field of multimodal scientific table understanding.

\bibliography{custom, zotero}

\newpage

\appendix
% \section{Appendix}

% \begin{table}[h]
% \centering
% \resizebox{0.79\linewidth}{!}
% {\begin{tabular}{lr}
%     \toprule
%     \textbf{Type} & \textbf{Value}\\
%     % \midrule
%     % \multicolumn{2}{c}{\textbf{\textit{Basic Insight}}} \\
%      \midrule
%     Tables & 3681 \\
%     Question Length(Avg) & 20.30 \\
%     Answer Length (Avg) & 8.52 \\
%     % \textbf{Columns Per Table} & 6.68 \\
%     % \textbf{Rows Per Table } & 16.71 \\
%     % \textbf{Ratio of Numerical Cells} & 65.74\% \\
%     Average Reasoning Steps & 6.26 \\
%     Training Set & 886 \\
%     Testing Set & 19,661 \\
%     \bottomrule
% \end{tabular} 
% }
% \caption{
% The reasoning types, the description of their subtypes, and their proportion in our dataset. 
% The number in parentheses is the proportion of each reasoning type.
% }
% \label{table:dataset_statistic}
% % \vspace{-15pt}
% \end{table}
\begin{table*}[ht]\footnotesize
\centering
% \small
\resizebox{0.99\linewidth}{!}{
\begin{tabular}{llcc}
\toprule
\textbf{Reasoning Type}  & \textbf{Description}  & \textbf{Avg. Reasoning Step} & \textbf{Prop.\%} \\
\midrule
% \multirow{2}{*}{Look Up ($4.7$)} 
  % & Table Look Up & Search for specific tables & $2.7$ \\
  % % & Text Look Up & Search for sentences in paragraphs & $6.1$ \\
  % % 查找文中某段文字
  % Arithmetic Calculation & Numerical calculations & $64.3$ \\
  Add & Calculate the sum between numbers &2.8 &21.1 \\
  Comparison & Comparison of values &2.1 & $13.7$ \\
  Domain Knowledge Calculation & Calculations need domain knowledge &2.2 & $1.5$ \\
  Divide & Perform division between numbers &3.4 &14.2 \\ 
  Look Up & Search for cells in tables &1.5 & $8.9$ \\
  Max/Min & Retrieve the maximum or minimum number &3.2 &15.7 \\
  Ranking & Arranges items in a specific order &2.4 & $9.6$ \\
  Subtract & Perform subtraction between numbers &4.1 & 15.3 \\
\bottomrule
\end{tabular}
}
\caption{
% 4种推理类型，及其子类的描述，以及在我们数据集中的比例
The reasoning types, descriptions, average reasoning step, and proportion in our dataset. 
% Question Category and Subcategories with Descriptions and Percentages
}
\label{tab:reasoning_type}
\end{table*}

% \begin{table}[ht]\footnotesize
% \centering
% % \small
% \resizebox{0.99\linewidth}{!}{
% \begin{tabular}{lcc}
% \toprule
% \textbf{Reasoning Type}   & \textbf{Avg. Reasoning Step} & \textbf{Prop.\%} \\
% \midrule

% Add  &2.8 &21.1 \\
% Comparison &2.1 & $13.7$ \\
% % Domain Knowledge   &2.2 & $1.5$ \\
% Divide  &3.4 &14.2 \\ 
% Look Up  &1.5 & $8.9$ \\
% Max/Min &3.2 &15.1 \\
% Ranking &2.4 & $9.6$ \\
% Subtract  &4.1 & 15.3 \\
% \bottomrule
% \end{tabular}
% }
% \caption{
% % 4种推理类型，及其子类的描述，以及在我们数据集中的比例
% The reasoning types, descriptions, and proportion in our MMSci-Ins and MMSci-Eval dataset. More details are provided in Appendix~\ref{app:detail} 
% % Question Category and Subcategories with Descriptions and Percentages
% }
% \label{tab:reasoning_type}
% \end{table}
\section{Details about MMSci}
\subsection{Datasets Statistics}
\label{app:detail}
Table~\ref{tab:reasoning_type} presents the distribution of reasoning types in our MMSci-Eval dataset. The most common type is addition (21.1\%), followed by subtraction (15.3\%) and max/min operations (15.7\%). Division and comparison operations also appear frequently (14.2\% and 13.7\% respectively). More complex operations like ranking (9.6\%) and look-up (8.9\%) occur less frequently, while domain knowledge calculations are rare (1.5\%).

The average number of reasoning steps varies significantly across types, with subtraction requiring the most steps (4.1) and look-up operations requiring the fewest (1.5). This variation reflects the inherent complexity of different mathematical operations and their application to tabular data. Notably, even seemingly simple operations like addition require multiple steps (2.8) on average, indicating the complexity of reasoning with tabular scientific data.

\begin{figure}[ht]
% \small
\centering
\includegraphics[width=0.99\linewidth]{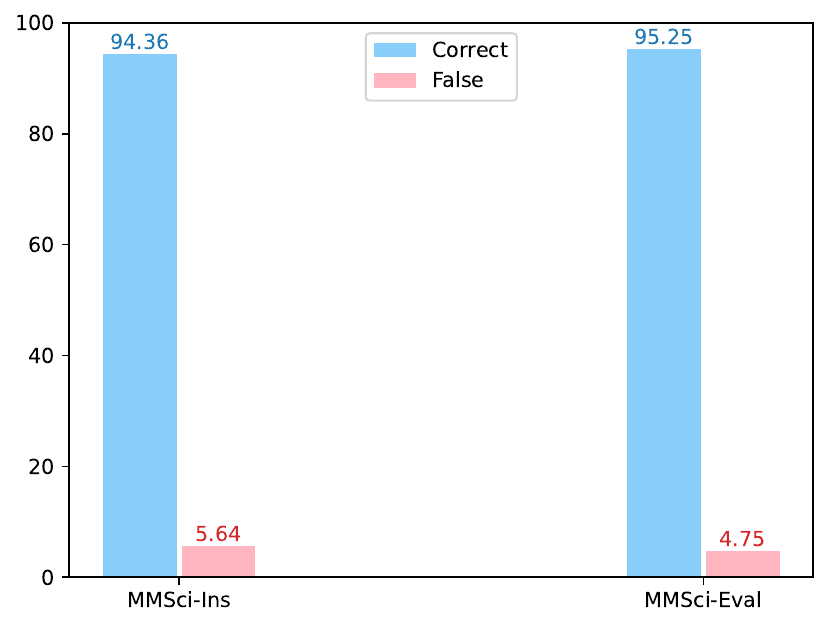}
\caption{Evaluation of generated data of MMSci-Ins and
MMSci-Eval dataset. Correct refers to the
data verified correctly by human annotators.}
\label{fig:MMSci-Pre example}
\end{figure}

\subsection{Dataset Quality Control }
\label{app:quality}

To ensure data quality, we conduct a rigorous human verification process for both MMSci-Ins and MMSci-Eval datasets. For MMSci-Ins, we manually verify 40\% of the generated samples, achieving a high accuracy rate of 94.36\%. For MMSci-Eval, given its critical role as a benchmark, we carefully examine all 3,114 generated samples and achieve an accuracy of 95.25\%. For any identified incorrect samples, we employ GPT-4o to regenerate them following the same self-consistency voting mechanism, followed by another round of both automatic and manual verification to ensure quality. This iterative process ensures the reliability and correctness of our datasets for both training and evaluation purposes.

%%%%%--------------\input{tables/consis}

\begin{table*}
    \centering
    \resizebox{0.79\linewidth}{!}{
    \begin{tabular}{lc|cc|cc}
    \toprule
     & \multicolumn{1}{c}{} & \multicolumn{2}{c}{\textbf{TQA}} & \multicolumn{2}{c}{\textbf{TFV}} \\ 
    % \cline{2-6}
      \multicolumn{1}{l}{\textbf{Model}} & \multicolumn{1}{c}{\textbf{Modal}}  & \multicolumn{1}{c}{\textbf{Acc.}}  & \multicolumn{1}{c}{\textbf{Consis.}}  & \multicolumn{1}{c}{\textbf{Acc.}}  & \multicolumn{1}{c}{\textbf{Consis.}}\\
    \hline
     \multirow{2}{*}{Qwen2-VL-7B-Ins.~\cite{wang2024qwen2}} & Text & \textbf{21.65} & \multirow{2}{*}{\textbf{60.40}} & \textbf{50.10} & \multirow{2}{*}{\textbf{72.48}} \\
     & Image &\textbf{39.11} & &\textbf{52.79} \\
     \hline
    % \cline{1-4}
    
    \multirow{2}{*}{LLaVA-NeXT-7B~\cite{li2024llavanext-strong}} & Text & 3.17 & \multirow{2}{*}{14.81} & 2.03  & \multirow{2}{*}{23.65} \\
    % \cdashline{2-3}
      & Image & 0.19&  & 0.86 \\
    \hline
    \multirow{2}{*}{MiniCPM-V-2.6-8B~\cite{yao2024minicpm}} & Text & 21.11 & \multirow{2}{*}{48.78} & 30.82 & \multirow{2}{*}{38.53} \\
    % \cdashline{2-3}
     & Image & 26.58  &  & 33.23  \\
    \hline
    \multirow{2}{*}{InternVL-2-8B~\cite{chen2024internvl}} & Text & 19.84 & \multirow{2}{*}{50.89}  & 42.87 & \multirow{2}{*}{36.42} \\
    % \cdashline{2-3}
      & Image &  25.72 &  & 44.99 \\
    \hline
    \multirow{2}{*}{Pixtral-12B~\cite{agrawal2024pixtral}} & Text & 1.44 & \multirow{2}{*}{16.52} & 4.43 & \multirow{2}{*}{29.88} \\
    % \cdashline{2-3}
    & Image &  0.96 & & 5.49 \\
    \hline
    % \cline{1-4} 
    % \multirow{2}{*}{LLaVA-NEXT-13B} & Text & 1.31 & \multirow{2}{*}{23.12}& 3.25 & \multirow{2}{*}{18.11} \\
    % \cdashline{2-3}
      % & Image &  2.31 & & 1.83 \\
    % \hline

    \multirow{2}{*}{Llama-3.2-11B-Vision-Ins.~\cite{meta2024llama3.2}} & Text & 3.24& \multirow{2}{*}{15.71} & 6.96 & \multirow{2}{*}{20.40}\\
    % \cdashline{2-3}
      & Image & 1.15 && 5.85 \\

    \bottomrule
\end{tabular}}
\caption{Vision-language consistency evaluation across different MLLMs. Consistency scores measure the percentage of identical responses between modalities, indicating the model's cross-modal alignment.}
\label{tab:consistency}
    \label{tab:main}
\end{table*}
%%%%%--------------\input{tables/consis}

\subsection{Prompt for Generating Data}
The prompt for MMSci-Ins and MMSci-Pre data generation is shown in Table~\ref{tab:prompt}.

\begin{table*}[ht]
\centering
\small
\begin{tabular}{p{0.9\textwidth}}
\toprule
\textbf{The prompt for Generating data} \\
\midrule
You are given a table image and a description: \{description\}.\\
1.For the Table to Text (T2T) task, come up with a one to two sentence length succinct multi-hop reasoning step of the description.
\\
Write your results as 'T2T Reasoning:' and then the succinct reasoning step.\\
\\
2.For the Table Question Answering (TQA) task, come up with a question and answer with multi-hop reasoning step. 

The question and answer must be based on the table image and description. 
\\\\

Write your results as 'TQA Question:' and then the question and 'TQA Reasoning:' and then the reasoning step and 'TQA Answer:' and then the answer.
\\
When generating 'TQA Question:', make sure it is a single question that requires reasoning based on the table.\\\\
When generating 'TQA Answer:', provide the final answer in the JSON structure, using the format {{"answer": "<YOUR ANSWER>"}}\\\\
Make sure the answer only contains one entity, such as 'So, the answer is {{"answer": "23"}}.'
\\\\
3.For the Table Fact Checking (TFV) task, come up with a statement and answer with multi-hop reasoning step.\\\\
The statement and answer must be based on the table image and description.
The table 'supports' or 'refutes' the statement. The statement should be considered 'not enough info' if it may or may not be true. \\
Write your results as 'TFV Statement:' and then the statement and 'TFV Reasoning:' and then the reasoning step and 'TFV Answer:' and then the answer.\\
Make sure the answer only contains one entity, such as 'Thus, the answer is {{"answer": "supports"}}.'\\\\

When generating 'TFV Answer:', provide the final answer in the JSON structure, using the format {{"answer": "<YOUR ANSWER>"}}
\\\\
Fill the result into JSON format without any other words: \\\\
    "T2T Reasoning": "<YOUR T2T REASONING>", \\
    "TQA Question": "<YOUR TQA QUESTION>", \\
    "TQA Reasoning": "<YOUR TQA REASONING>", \\
    "TQA Answer": "<YOUR TQA ANSWER>",\\
    "TFV Statement": "<YOUR TFV STATEMENT>", \\
    "TFV Reasoning": "<YOUR TFV REASONING>", \\
    "TFV Answer": "<YOUR TFV ANSWER>"\\\\

Examples:\\

\{TQA Examples\}\\
\{TFV Examples\}\\
\{T2T Examples\}\\
\bottomrule
\end{tabular}
\caption{
The prompts for generating the questions, reasoning steps, and answers or claims of MMSci-Ins and MMSci-Eval datasets.
}
\label{tab:prompt}
\end{table*}

\section{Experimental Settings}
\noindent\textbf{Implementation Details.} Both models follow a three-component design. Qwen2-VL-7B-Instruct consists of a Vision Transformer (ViT)~\cite{dosovitskiy2020image} as the vision tower, a MLP as the vision-language connector, and Qwen2~\cite{wang2024qwen2} as the language model. LLaVA-NeXT-7B uses a pre-trained CLIP model~\cite{clip_paper} as the visual encoder, a MLP connector, and Vicuna-7B~\cite{vicuna2023} as the backbone. In both architectures, the vision encoder processes images into visual features, which are projected into the LLM's word embedding space via the MLP connector.

\noindent\textbf{Training Details.} All experiments are conducted on 4$\times$A100 80GB GPUs using LoRA with rank 64 and sequence length 4096. For table structure learning, LLaVA-NeXT-7B requires 15 hours for MMTab-Pre (150k), 3 hours for MMSci-Pre (52k), and 20 hours for combined training (one epoch). Qwen2-VL-7B takes 15 hours, 8 hours, and 19 hours respectively. The instruction tuning stage requires approximately 1 hour for 4 epochs with 12k samples for both models.

\section{More Experimental Results and Analysis}

\subsection{Vision-Language Consistency Analysis} \label{sec:consist}
We evaluate the cross-modal consistency of different MLLMs by comparing their performance when processing table information through different modalities. For each model, we test with both table images (image modality) and their textual representations (text modality), measuring both task performance (Acc.) and cross-modal consistency (Consis.).

Qwen2-VL-7B-Instruct demonstrates superior cross-modal alignment, achieving the highest consistency scores on both TQA (60.40\%) and TFV (72.48\%) tasks. Notably, it maintains strong performance across both modalities, with image-based accuracy (TQA: 39.11\%, TFV: 52.79\%) consistently outperforming text-based results (TQA: 21.65\%, TFV: 50.10\%). This suggests robust integration of visual and textual understanding capabilities.

Other models show varying degrees of modality gap. MiniCPM-V-2.6-8B and InternVL-2-8B achieve moderate consistency (48.78\% and 50.89\% on TQA), while models like LLaVA-NeXT-7B and Pixtral-12B show significant disparities between modalities, resulting in lower consistency scores. These results highlight the challenge of maintaining consistent reasoning capabilities across different input modalities in table understanding tasks.

\section{Dataset Examples and Case Study}
\subsection{Dataset Examples}
The training examples of MMSci-Pre Dataset are shown
in Figure~\ref{fig:MMSci-Pre example}. The examples of MMSci-Eval are shown  Figure~\ref{fig:MMSci-ins case1}.
\subsection{Case Study}
\begin{figure*}[ht]
% \small
\centering
\includegraphics[width=0.99\linewidth]{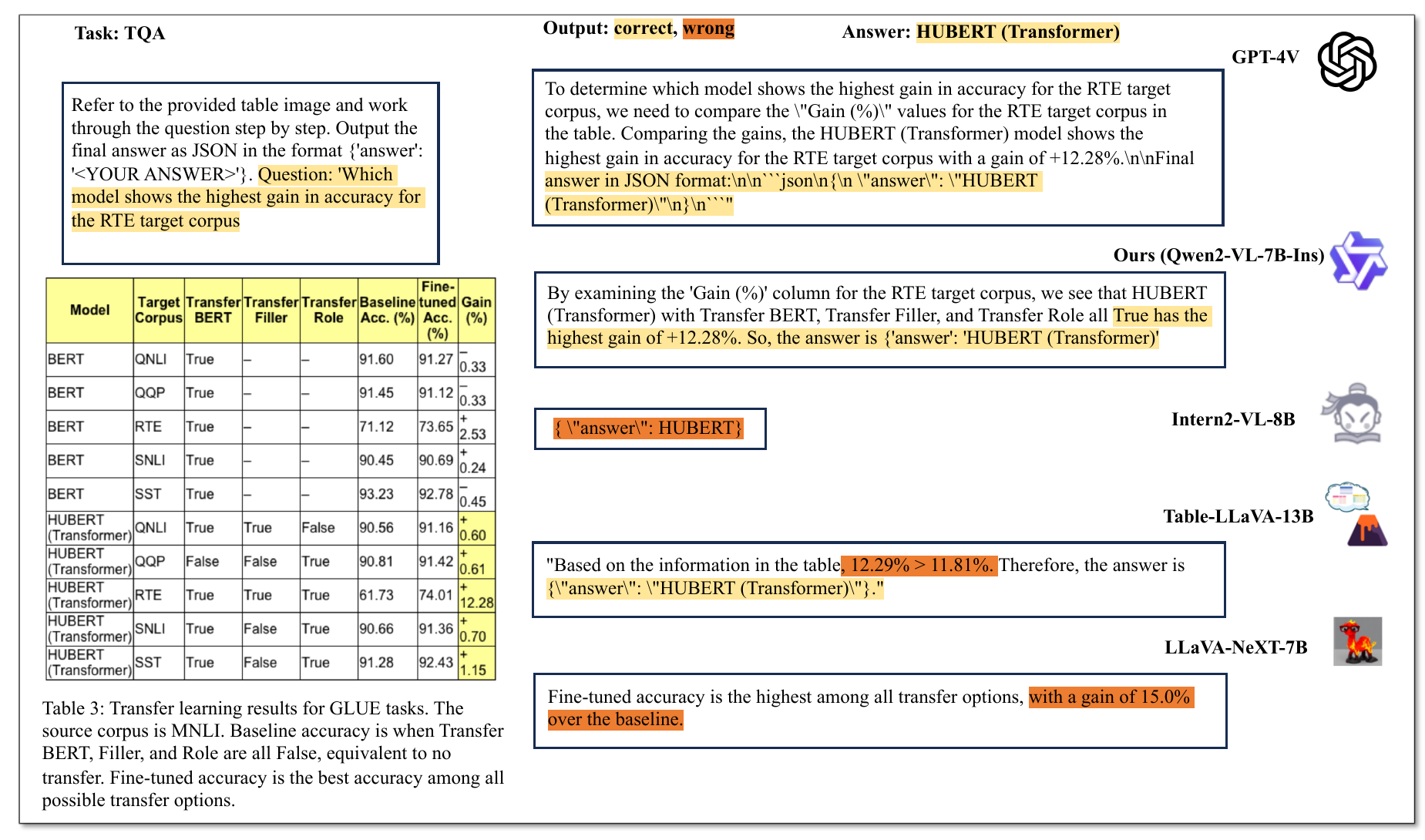}
\caption{A case example of TQA task
}
\label{fig:MMSci-ins case1}
\end{figure*}

\begin{figure*}[ht]
% \small
\centering
\includegraphics[width=0.99\linewidth]{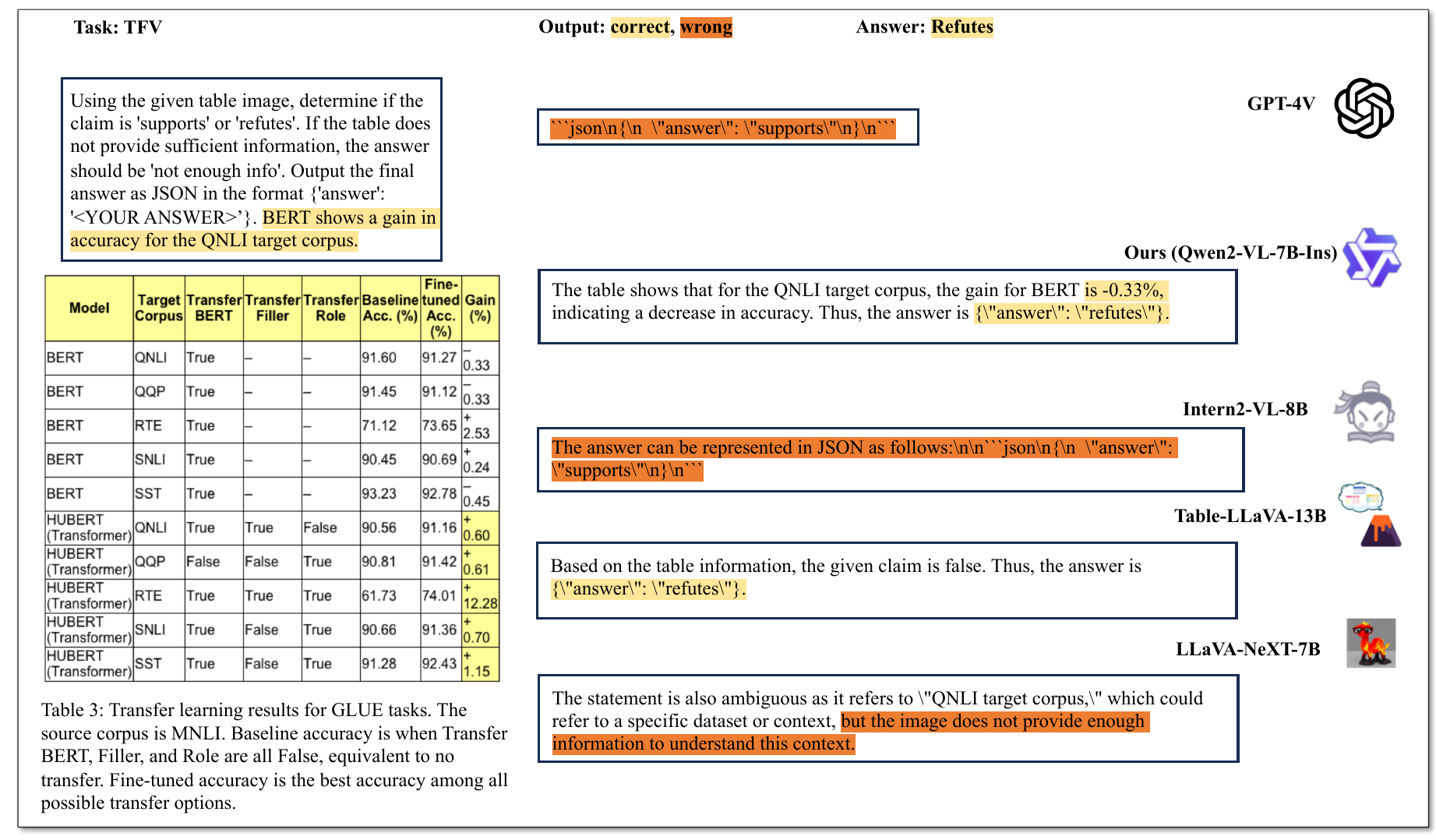}
\caption{A case example of TFV task
}
\label{fig:MMSci-ins case2}
\end{figure*}

The case study of TQA and TFV tasks are shown in Figure~\ref{fig:MMSci-ins case1} and Figure~\ref{fig:MMSci-ins case2}, respectively

\begin{figure*}[ht]
% \small
\centering
\includegraphics[width=0.99\linewidth]{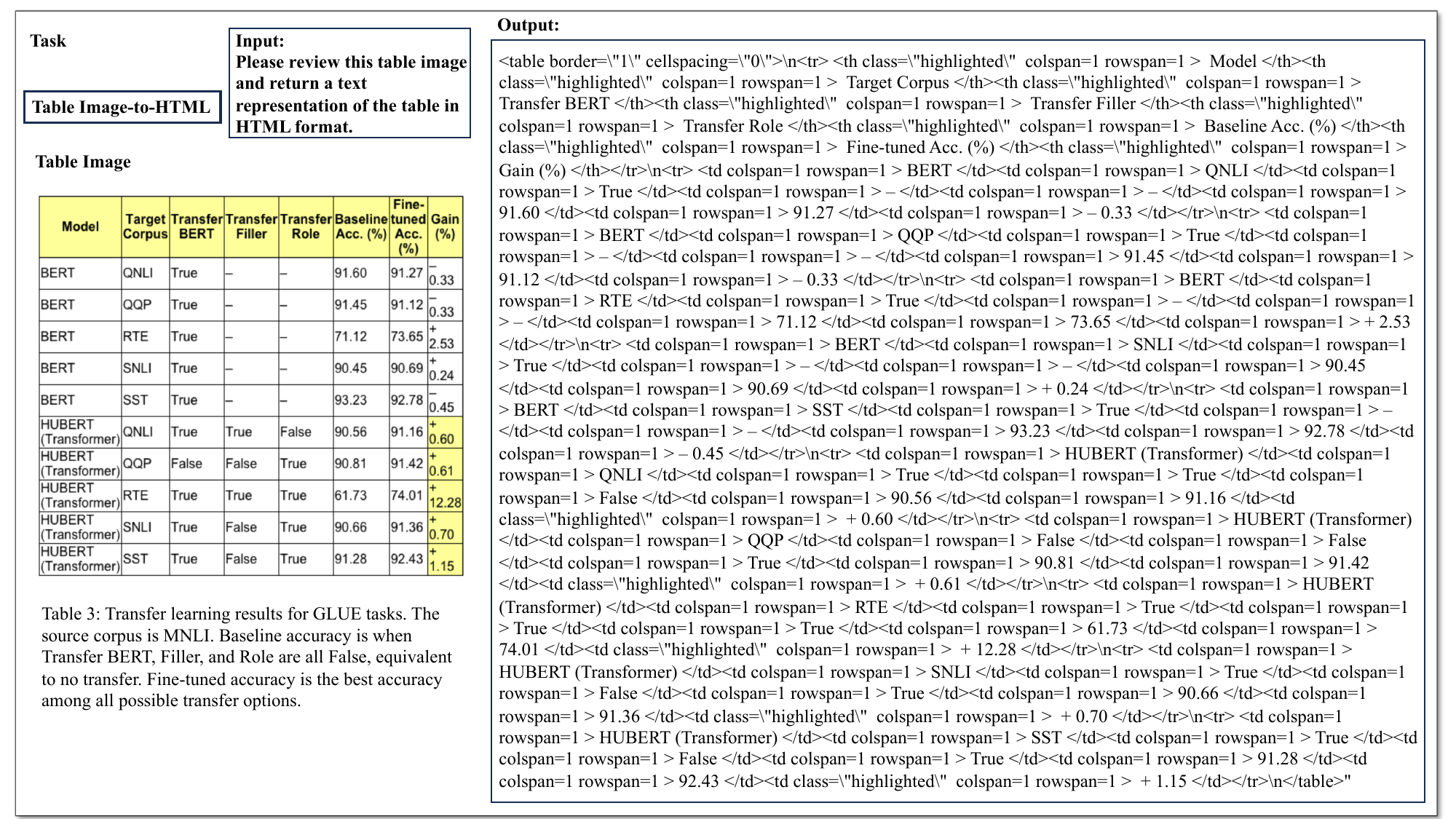}
\caption{MMSci-Pre Dataset example
}
\label{fig:MMSci-Pre example}
\end{figure*}

\begin{figure*}[ht]
% \small
\centering
\includegraphics[width=0.99\linewidth]{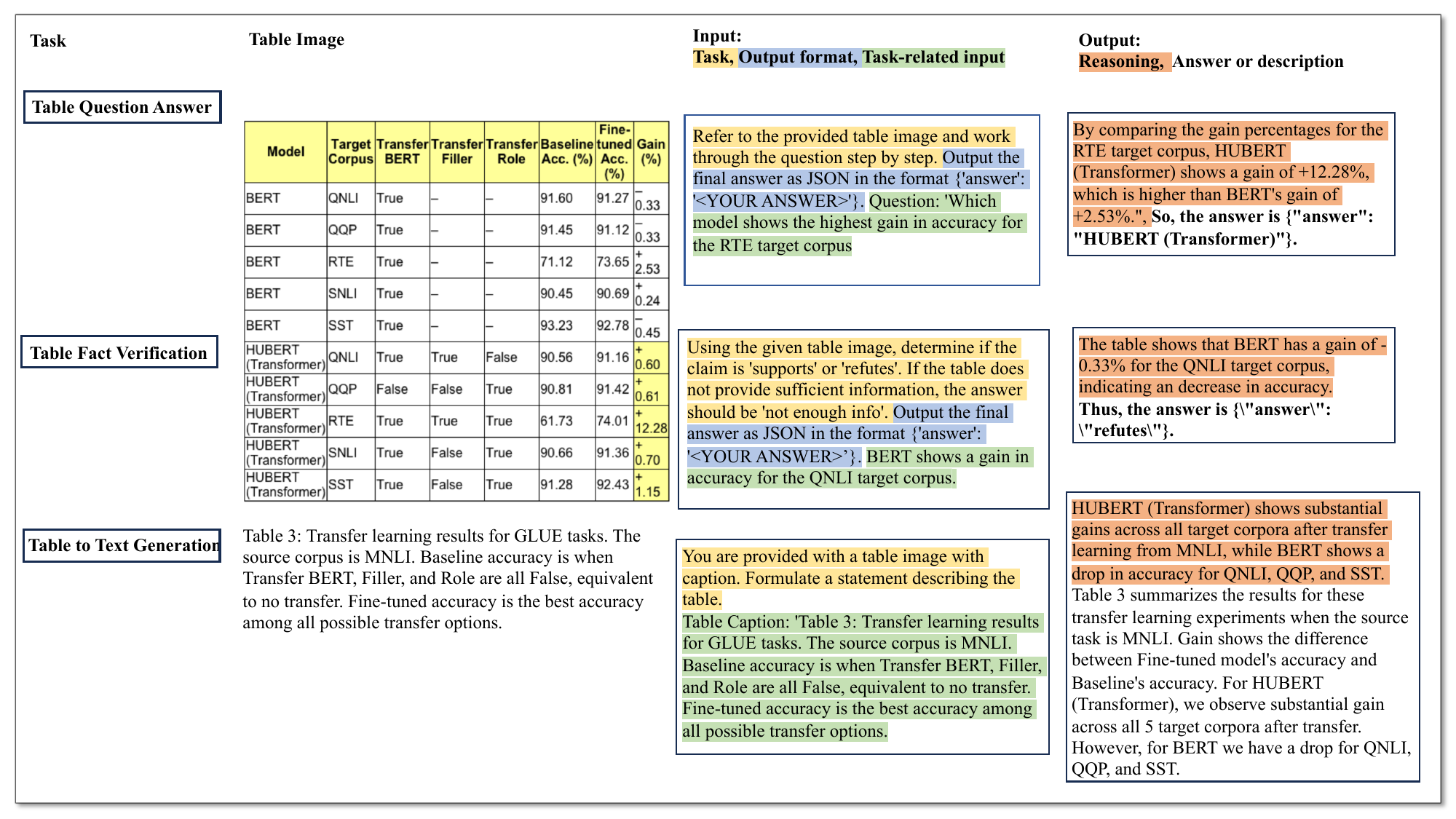}
\caption{MMSci-Ins and MMSci-Eval Dataset example
}
\label{fig:MMSci-ins example1}
\end{figure*}

\end{document}